\documentclass[11pt]{article}
\usepackage[dvips]{graphicx}
\usepackage{amssymb,amsmath,color,mathtools}
\usepackage{url}

\usepackage[mathcal]{eucal}

\DeclareMathAlphabet\mathbfcal{OMS}{cmsy}{b}{n}

\newcommand{\BEAS}{\begin{eqnarray*}}
\newcommand{\EEAS}{\end{eqnarray*}}
\newcommand{\BEA}{\begin{eqnarray}}
\newcommand{\EEA}{\end{eqnarray}}
\newcommand{\BEQ}{\begin{equation}}
\newcommand{\EEQ}{\end{equation}}
\newcommand{\BIT}{\begin{itemize}}
\newcommand{\EIT}{\end{itemize}}
\newcommand{\rb  }{\mathbb{R}}
\newcommand{\E}{\mathbb{E}}
\usepackage{subfigure}

\newcommand{\mysec}[1]{Section~\ref{sec:#1}}
\newcommand{\eq}[1]{Eq.~(\ref{eq:#1})}
\newcommand{\myfig}[1]{Figure~\ref{fig:#1}}

\newtheorem{theorem}{Theorem}

\numberwithin{equation}{section}

\begin{document}

\title{Gradient Descent on Infinitely Wide Neural Networks: \\ Global Convergence and Generalization }

\author{
  Francis Bach \\
  Inria \& Ecole Normale Sup\'erieure \\
  PSL Research University  \\
\texttt{francis.bach@inria.fr} \\
 \and
 L\'ena\"ic Chizat \\
 Ecole Polytechnique F\'ed\'erale de Lausanne \\ \texttt{lenaic.chizat@epfl.ch}
}

\maketitle

\begin{abstract}
Many supervised machine learning methods are naturally cast as optimization problems. For prediction models which are linear in their parameters, this often leads to convex problems for which many mathematical guarantees exist. Models which are non-linear in their parameters such as neural networks lead to non-convex optimization problems for which guarantees are harder to obtain. In this review paper, we  consider two-layer neural networks with homogeneous activation functions where the number of hidden neurons tends to infinity, and show how qualitative convergence guarantees may be derived. 
\end{abstract}


\section{Introduction}

In the past twenty years, data in all their forms have played an increasing role: in personal lives, with various forms of multimedia and social networks, in the economic sector where most industries monitor all of their processes and aim at making data-driven decisions, and in sciences, where data-based research is having more and more impact, both in fields which are traditionally data-driven such as medecine and biology, but also in humanities.

This proliferation of data leads to a need for automatic processing, with striking recent progress in some perception tasks where humans excel, such as image recognition, or natural language processing. These advances in artificial intelligence were fueled by the combination of three factors: (1) massive data to learn from, such as millions of labeled images, (2) increased computing resources to treat this data, and (3) continued scientific progress in algorithms.

Machine learning is one of the scientific disciplines that have made this progress possible, by blending statistics and optimization to design algorithms with theoretical generalization guarantees. The goal of this review paper  that will be published in the Proceedings of the 2022 International Congress of Mathematicians is to highlight our recent progress from already published works~\cite{chizat2018global,chizat2020implicit}, and to present a few open mathematical problems.

 \section{Supervised learning}
 In this paper, we will focus on the supervised machine learning problem, where we are being given $n$ pairs of observations 
$(x_i,y_i) \in \mathcal{X} \times \mathcal{Y}$, $i=1,\dots,n$, for example images ($\mathcal{X}$ is then the set of all possible images), with a set of labels ($\mathcal{Y}$ is then a finite set, which we will assume to be a subset of $\rb$ for simplicity).
The goal is to be able to predict a new output $y \in \mathcal{Y}$, given a previously unobserved input $x \in \mathcal{X}$.

Following the traditional statistical \emph{M-estimation} framework~\cite{van2000asymptotic}, this can be performed by considering prediction functions 
$x\mapsto h(x,\theta) \in \rb$, parameterized by $\theta \in   \rb^d$. 
The vector $\theta$ is then estimated through regularized empirical risk minimization, that is by solving
\BEQ
\label{eq:erm}
\min_{\theta \in \rb^d}\   \  \frac{1}{n} \sum_{i=1}^n  \ell \big(y_i,h(x_i,\theta) \big)   +   \lambda  \Omega(\theta)   ,
\EEQ
 where $\ell: \mathcal{Y} \times \rb \to \rb$ is a loss function, and $\Omega: \rb^d \to \rb$ is a regularization term that avoids overfitting (that is learning a carbon copy of the observed data that does not generalize well to unseen data).
 
Typical loss functions are the square loss   $\ell\big(y_i,h(x_i,\theta) \big) = \frac{1}{2} \big(y_i - h(x_i,\theta) \big)^2$ for regression problems, and the logistic loss   $\ell\big(y_i,h(x_i,\theta) \big)  = \log \big( 1 + \exp( - y_i h(x_i,\theta) )\big)$ for binary classification where $\mathcal{Y} = \{-1,1\}$. In this paper, we will always assume that the loss function is continuously twice differentiable and convex with respect to the second variable. This applies to a wide variety of output spaces beyond regression and binary classification (see~\cite{nowak2019general} and references therein).

When the predictor depends linearly in the parameters, typical regularizers are the squared Euclidean norm $\Omega(\theta) = \frac{1}{2} \| \theta\|_2^2$ or the $\ell_1$-norm $\Omega(\theta) = \| \theta\|_1$, that both lead to improved generalization performance, with the $\ell_1$-norm providing additional variable selection benefits~\cite{giraud2021introduction}.

\subsection{Statistics and optimization} The optimization problem in \eq{erm} leads naturally to two sets of questions, which are often treated separately. Given that some minimizer $\hat{\theta}$ is obtained (no matter how), how does the corresponding prediction function generalize to unseen data? This is a statistical question that requires assumptions on the link between the observed data (usually called the ``training data''), and the unseen data  (usually called the ``testing data''). It is typical to assume that the training and testing data are sampled independently and identically from the same fixed distribution. Then a series of theoretical guarantees applies, based on various probabilistic concentration inequalities~(see, e.g.,~\cite{mohri2018foundations}).

The second question is how to obtain an approximate minimizer $\hat{\theta}$, which is an optimization problem, regardless on the relevance of $\hat{\theta}$ on unseen data~(see, e.g., \cite{bubeck2015convex}). For high-dimensional problems where $d$ is large (up to millions or billions),  classical gradient-based algorithms are preferred because of their simplicity, efficiency, robustness and favorable convergence properties. The most classical one is gradient descent, which is an iterative algorithm with iteration:
$$
\theta_k = \theta_{k-1} - \gamma \nabla \mathcal{R}(\theta_{k-1}),
$$
where $\mathcal{R}(\theta) =  \frac{1}{n} \sum_{i=1}^n  \ell \big(y_i,h(x_i,\theta) \big)   +   \lambda  \Omega(\theta) $ is the objective function in \eq{erm}, and $\gamma> 0$ the step-size.

In this paper, where we aim at tackling high-dimensional problems, we will often consider the two problems of optimization and statistical estimation jointly.

\subsection{Linear predictors and convex optimization}
\label{sec:cvx}
In many applications, a prediction function which is linear in the parameter $\theta$ is sufficient for good predictive performance, that is, we can write $$h(x,\theta) = \theta^\top \Phi(x)$$ for some function $\Phi: \mathcal{X} \to \rb^d$, which is often called a ``feature function''. For simplicity we have assumed finite-dimensional features, but infinite-dimensional features can also be considered, with a specific computational argument to allow finite-dimensional computations through reproducing kernel Hilbert spaces (see, e.g.,~\cite{scholkopf-smola-book} and references therein).

 Given a convex loss function, the optimization problem is convex and gradient descent on the objective function, together with its stochastic extensions, has led to a number of efficient algorithms with strong generalization guarantees of convergence towards the \emph{global} optimum of the objective function~\cite{bubeck2015convex}. For example, for the square loss or the logistic loss, if the feature function is bounded in $\ell_2$-norm by $R$ for all observations, and for the  squared Euclidean norm $\Omega(\theta) = \frac{1}{2} \| \theta\|_2^2$, bounds on the number of iterations to reach a certain precision $\varepsilon$ (difference between the candidate function value $\mathcal{R}(\theta)$ and the minimal value) can be obtained:
 \BIT
 \item For gradient descent, $\frac{R^2}{\lambda} \log \frac{1}{\varepsilon}$ iterations are needed, but each iteration has a running time complexity of $O(nd)$, because the $d$-dimensional gradients of the $n$ functions $\theta \mapsto  \ell \big(y_i,h(x_i,\theta) \big) $, $i=1,\dots,n$, are needed.
 \item For stochastic gradient descent, with iteration $\theta_k = \theta_{k-1} - \gamma \nabla \ell \big(y_{i(k)},h(x_{i(k)},\theta_{k-1}) \big) $, with $i(k) \in \{1,\dots,n\}$ taken uniformly at random, the number of iterations is at most  $\frac{R^2}{\lambda}  \frac{1}{\varepsilon}$. We lose the logarithmic dependence, but each iteration has complexity $O(d)$, which can be a substantial gain when $n$ is large.
 \item More recent algorithms based on variance reduction can achieve an overall complexity proportional to $\big( n + \frac{R^2}{\lambda} \big) \log \frac{1}{\varepsilon}$, thus with an exponential convergence rate at low iteration cost (see~\cite{gower2020variance} and references therein).
 
 \EIT
 
 In summary, for linear models, algorithms come with strong performance guarantees that reasonably match their empirical behavior. As shown below, non-linear models exhibit more difficulties.

\subsection{Neural networks and non-convex optimization}
\begin{figure}
\begin{center}
\includegraphics[scale=.95]{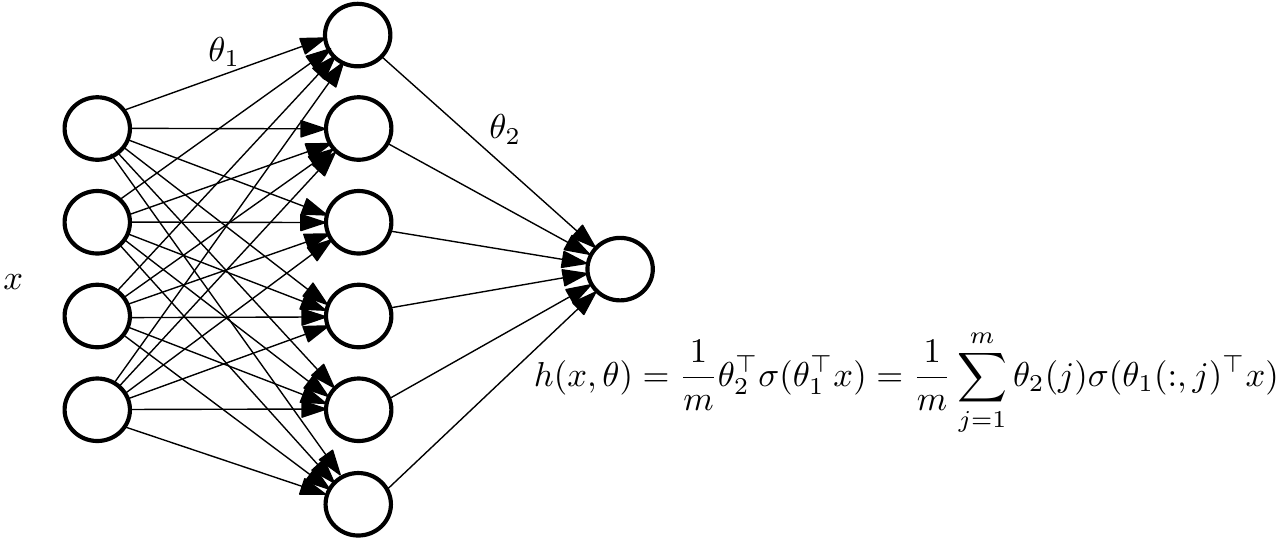}

 \vspace*{-.5cm}

\end{center}
\caption{Neural network with a single hidden layer, with an input weight matrix $\theta_1 \in \rb^{d \times m}$ and a output weight vector $\theta_2 \in \rb^{m}$.}
\label{fig:nn}
\end{figure}

In many other application areas, in particular in multimedia processing, linear predictors have been superseded by non-linear predictors, with neural networks being the most classical example (see~\cite{goodfellow2016deep}). A vanilla neural network is a prediction function of the form 
$$ h(x,\theta) =   \theta_s ^\top\sigma ( \theta_{s-1}^\top \sigma( \cdots\theta_2^\top \sigma (\theta_1^\top x) ), $$
where the function $\sigma: \rb \to \rb$ is taken component-wise, with the classical examples being the sigmoid function $\sigma(t) = ( 1 + \exp(-t))^{-1}$ and the ``rectified linear unit'' (ReLU), $\sigma(t) = t_+ = \max\{t,0\}$. The matrices $\theta_1,\dots,\theta_s$ are called weight matrices. The simplest non-linear predictor is for $s=2$, and will be the main subject of study in this paper. See \myfig{nn} for an illustration.

The main difficulty is that now the optimization problem in \eq{erm} is not convex anymore, and gradient descent can converge to stationary points that are not global minima. Theoretical guarantees can be obtained regarding the decay of the norm of the gradient of the objective function, or convergence to a local minimizer may be ensured~\cite{lee2016gradient,jin2017escape}, but this does not exclude bad local minima, and global quantitative convergence guarantees can only be obtained with exponential dependence in dimension for the class of (potentially non-convex) functions of a given regularity~\cite{nesterov2018lectures}.

An extra difficulty is related to the number of hidden neurons, also referred to as the \emph{width} of the network (equal to the size of $\theta_2$ when $s=2$), which is often very large in practice, which poses both statistical and optimization issues. We will see that this is precisely this overparameterization that allows to obtain qualitative  global convergence guarantees.

\section{Mean field limit of overparameterized one-hidden layer neural networks}
We now tackle the study of neural networks with one infinitely wide hidden layer. They are also referred to as (wide) two-layer neural networks, because they have two layers of weights. We first rescale  the prediction function by $1/m$ (which can be obtained by rescaling $\theta_2$ by $1/m$), and express it explicitly as en empirical average,  as 
$$
  h(x,\theta) =  \frac{1}{m} \theta_2^\top \sigma( x^\top \theta_1 )
= \frac{1}{m} \sum_{j=1}^m \theta_2(j) \cdot \sigma\big[x^\top  \theta_1(\cdot,j) \big],
$$
where $\theta_2(j) \in \rb$ is the output weight associated to neuron $j$, and $\theta_1(\cdot,j) \in \rb^d$ the corresponding vector of input weights. The key observation is that the prediction function $x \mapsto h(x,\theta)$ is the average of $m$ prediction functions
$x \mapsto \theta_2(j) \cdot \sigma\big[ \theta_1(\cdot,j)^\top x\big]$, for $j=1,\dots,m$, with \emph{no sharing of the parameters} (which is not true if extra layers of hidden neurons are added).

In order to highlight this parameter separability, we define 
$$w_j = \big[ \theta_2(j) ,  \theta_1(\cdot,j)\big] \in \rb^{d+1}$$ the set of weights associated to the hidden neuron $j$, and we define
$$
\Psi(w): x \mapsto w{(1)} \cdot \sigma\big[ x^\top w{(2,\dots,d+1)}\big],
$$
so that the prediction function $x \mapsto h(\cdot,w_1,\dots,w_m)$, parameterized by $w_1,\dots,w_m$, is now
\BEQ
\label{eq:hemp}
h(\cdot,w_1,\dots,w_m) = \frac{1}{m} \sum_{j=1}^m \Psi(w_j).
\EEQ

The empirical risk is of the form $$R(h) = \E \big[ \ell( y, h(x)) \big] ,$$ which is convex in $h$ for convex loss functions (even for neural networks), but typically non convex in $w$. Note that the resulting problem of minimizing a convex function $R(h)$ for $h = \frac{1}{m} \sum_{j=1}^m \Psi(w_j)$ applies beyond neural networks, for example, for sparse deconvolution~\cite{chizat2021sparse}.

\subsection{Reformulation with probability measures}
We now define by $\mathcal{P}(\mathcal{W})$ the set of probability measures on $\mathcal{W}=\rb^{d+1}$. We can rewrite \eq{hemp} as $$
h = \int_\mathcal{W} \Psi(w) d\mu(w),
$$
with $\mu = \frac{1}{m} \sum_{j=1}^m \delta_{w_j}$ an average of Dirac measures at each $w_1,\dots,w_m$. Following a physics analogy, we will refer to each $w_j$ as a \emph{particle}. When the number $m$ of particles grow, then the empirical measure $\frac{1}{m} \sum_{j=1}^m \delta_{w_j}$ may converge in distribution to a probability measure with a density, often referred to as a \emph{mean field} limit. Our main reformulation will thus be to consider an optimization problem over probability measures.

The optimization problem we are faced with is equivalent to
\BEQ
\inf_{ \mu \in \mathcal{P}(\mathcal{W})} R \Big(  \int_\mathcal{W}\Psi(w) d\mu(w) \Big),
\EEQ
with the constraint that $ \mu$ is an average of $m$ Dirac measures. In this paper, following a long line of work in statistics and signal processing~\cite{barron1993universal,kurkova}, we consider the optimization problem \emph{without this constraint}, and relate optimization algorithms for finite but large $m$ (thus acting on $W = (w_1,\dots,w_m)$ in $\mathcal{W}^m$) to a well-defined algorithm in $ \mathcal{P}(\mathcal{W})$.

Note that we now have a convex optimization problem, with a convex objective in $\mu$ over a convex set (all probability measures). However, it is still an infinite-dimensional space that requires dedicated finite-dimensional algorithms. In this paper we focus on gradient descent on $w$, which corresponds to standard practice in neural networks (e.g., back-propagation). For algorithms based on classical convex optimization algorithms such as the Frank-Wolfe algorithm, see~\cite{bach2017breaking}.

\subsection{From gradient descent to gradient flow}
Our general goal is to study the gradient descent recursion on $W = (w_1,\dots,w_m) \in \mathcal{W}^m$, defined as
\BEQ
\label{eq:gd}
W_{k} = W_{k-1} - \gamma m \nabla G(W_{k-1})
,\EEQ
with $$G(W) = R\big(h(\cdot,w_1,\dots,w_m) \big) = R \Big(  \frac{1}{m} \sum_{j=1}^m \Psi(w_j)\Big).$$
In the context of neural networks, this is exactly the back-propagation algorithm. We include the factor $m$ in the step-size to obtain a well-defined limit when $m$ tends to infinity (see \mysec{wass}).

For convenience in the analysis, we look at the limit when the step-size $\gamma$ goes to zero. If we consider a function $V: \rb \to  \mathcal{W}^m$, with values $V(k\gamma) = W_k$ at $t=k\gamma$, and we interpolate linearly between these points, then, we obtain exactly the standard Euler discretization of the ordinary differential equation (ODE)~\cite{suli2003introduction}:
\BEQ
\label{eq:gf}
\dot{V} = - m \nabla G(V).
\EEQ
 This gradient flow will be our main focus in this paper. As highlighted above, and with extra regularity assumptions, it is the limit of the gradient recursion in \eq{gd} for vanishing step-sizes $\gamma$. Moreover, under appropriate conditions, stochastic gradient descent, where we only observe an unbiased noisy version of the gradient, also leads in the limit $\gamma \to 0$ to the same ODE~\cite{kushner:yin:2003}. This allows to apply our results to probability distributions of the data $(x,y)$ which are not the observed empirical distribution, but the unseen test distribution, where the stochastic gradients come from the gradient of the loss from a single observation.

Three questions now emerge:
\begin{enumerate}
\item What is the  limit (if any) of the gradient flow in \eq{gf} when the number of particles $m$ gets large?
\item Where can the gradient flow converge to?
\item Can we ensure a good generalization performance when the number of parameters grows unbounded?
\end{enumerate}
In this paper, we will focus primarily in the next sections on the first two questions, and  tackle the third question in \mysec{generalization}.

\subsection{Wasserstein gradient flow}
\label{sec:wass}
Above, we have described a general framework where  we want to minimize a function $F$ defined on probability measures:
\BEQ
\label{eq:fmu}
F(\mu) = \displaystyle R \Big(  \int_{\mathcal{W}} \Psi(w) d\mu(w) \Big), 
\EEQ
with an algorithm  minimizing $\displaystyle G(w_1,\dots,w_m) = 
R \Big( \frac{1}{m}\sum_{j=1}^m \Psi(w_j) \Big)  $
through the gradient flow $\dot{W} = - m \nabla G(W)$, with $W = (w_1,\dots,w_m)$.


As shown in a series of works concerned with the infinite width limit of two-layer neural networks~\cite{nitanda2017stochastic,chizat2018global,mei2018mean,sirignano2020mean,rotskoff2018parameters}, this converges to a well-defined mathematical object called a Wasserstein gradient flow~\cite{ambrosio2008gradient}. This is a gradient flow derived from the Wasserstein metric on the set of probability measures, which is defined as~\cite{santambrogio2015optimal}, 
$$
W_2(\mu,\nu) ^2= \inf_{\gamma \in \Pi(\mu,\nu)} \int \| v-w\|_2^2 d\gamma(v,w),
$$
where $\Pi(\mu,\nu)$ is the set of probability measures on $\mathcal{W} \times \mathcal{W} $ with marginals $\mu$ and $\nu$. In a nutshell, the gradient flow is defined as the limit when $\gamma$ tends to zero of the extension of the following discrete time dynamics:
$$
\mu(t+\gamma) = \inf_{ \nu \in \mathcal{P}(\mathcal{W})} F(\nu) + \frac{1}{2 \gamma} W_2(\mu(t),\nu) ^2.
$$
When applying such a definition in a Euclidean space with the Euclidean metric, we recover the usual gradient flow $\dot{\mu} = - \nabla F(\mu)$, but here with the Wasserstein metric, this defines a specific flow on the set of measures. 
When the initial measure is a weighted sum of Diracs, this is exactly asymptotically (when $\gamma \to 0$) equivalent to backpropagation. When initialized with an arbitrary probability measure, we obtain a partial differential equation (PDE), satisfied in the sense of distributions.
Moreover, when the sum of Diracs converges in distribution to some measure, the flow converges to the solution of the PDE. More precisely, assuming $\Psi: \rb^{d+1} \to \mathcal{F}$ a Hilbert space, and $\nabla R(h) \in \mathcal{F}$ the gradient of~$R$, we consider the \emph{mean potential}
\BEQ
\label{eq:J}
J(w | \mu) = \Big\langle \Psi(w), \nabla R\Big( \int_{\mathcal{W}} \Psi(v) d\mu(v) \Big) \Big\rangle.
\EEQ
 The PDE is then the classical continuity equation:
\BEQ
\label{eq:mu}
\partial_t \mu_t(w) = {\mathrm{div}} ( \mu_t(w) \nabla J(w | \mu_t ) ),
\EEQ
which is understood in the sense of distributions. The following result formalizes this behavior~(see \cite{chizat2018global} for details and a more general statement).

\begin{theorem}
Assume that $R:\mathcal{F} \to [0,+\infty[$ and $\Psi:\mathcal{W}=\rb^{d+1} \to \mathcal{F}$ are (Fréchet) differentiable with Lipschitz differentials, and that $R$ is Lipschitz on its sublevel sets. Consider a sequence of initial weights $(w_j(0))_{j\geq 1}$ contained in a compact subset of $\mathcal{W}$ and let $\mu_{t,m} \coloneqq \frac1m \sum_{j=1}^m w_j(t)$ where $(w_1(t),\dots,w_m(t))$ solves the ODE~\eqref{eq:gf}. If $\mu_{0,m}$ weakly converges to some $\mu_0 \in \mathcal{P}(\mathcal{W})$ then $\mu_{t,m}$ weakly converges to $\mu_t$ where $(\mu_t)_{t\geq 0}$ is the unique weakly continuous solution to~\eqref{eq:mu} initialized with $\mu_0$.
\end{theorem}

In the following section, we will study the solution of this PDE (i.e., the Wasserstein gradient flow), interpreting it as the limit of the gradient flow in \eq{gf}, when the number of particles $m$ tend to infinity.

\section{Global convergence}
We consider the Wasserstein gradient flow defined above, which leads to the PDE in \eq{mu}. Our goal is to understand when we can expect that when $t \to \infty$, $\mu_t$ converges to a global minimum of $F$ defined in \eq{fmu}. Obtaining a global convergence result is not out of the question because $F$ is a convex functional defined on the convex set of probability measures. However, it is non trivial because with our choice of the Wasserstein geometry on measures, which allows an approximation through particles,  the flow has some stationary points which are not the global optimum (see examples in \mysec{experr}).

We start with an informal general result without technical assumptions before stating a formal simplified result.

\subsection{Informal result}
In order to avoid too many technicalities, we first consider an informal theorem in this paper and refer to~\cite{chizat2018global} for a detailed set of technical assumptions (in particular smoothness assumptions). This leads to the informal theorem:

\begin{theorem}[Informal]
If the support of the initial distribution includes all directions in $\rb^{d+1}$, and if the function $\Psi$ is positively $2$-homogeneous then if the  Wasserstein gradient flow weakly converges to a distribution, it can only be to a global optimum of $F$.
\end{theorem}

In~\cite{chizat2018global} another version of this result that allows for \emph{partial} homogeneity (e.g.,~with respect to a subset of variables) of degree $1$ is proven, at the cost of a more technical assumption on the initialization. For neural networks, we have $\Psi(w_j)(x) = m \theta_2(j) \cdot \sigma\big[ \theta_1(\cdot,j)^\top x \big]$, and this more general version applies. For the classical ReLU activation function $u \mapsto \max\{0,u\}$, we get a positively $2$-homogeneous function, as required in the previous statement.
A simple way to spread all directions is to initialize neural network weights from Gaussian distributions, which is standard in applications~\cite{goodfellow2016deep}.

\paragraph{From qualitative to quantitative results?}
Our result states that for infinitely many particles, we can only converge to a global optimum (note that we cannot show that the flow always converges). However, it is only a qualitative result in comparison with what is known for convex optimization problems in \mysec{cvx}:
\BIT

\item This is only for $m=+\infty$, and we cannot provide an estimation of the number of particles needed to approximate the mean field regime that is not exponential in $t$ (see such results e.g.~in~\cite{mei2019mean}). 

\item We cannot provide an estimation of the performance as the function of time, that would provide an upper bound on the running time complexity.

\EIT

Moreover, our result does not apply beyond a single hidden layer, and understanding the non-linear infinite width limits for deeper networks is an important research area~\cite{nguyen2020rigorous,araujo2019mean,fang2021modeling,hanin2019finite,sirignano2021mean,wojtowytsch2020banach,yang2020feature}.

\paragraph{From informal to formal results.} Beyond the lack of quantitative guarantees, obtaining a formal result requires regularity and compactness assumptions which are not satisfied for the classical ReLU activation function $u \mapsto \max\{0,u\}$, which is not differentiable at zero (a similar result can be obtained in this case but under stronger assumptions on the data distribution and the initialization~\cite{wojtowytsch2020convergence,chizat2020implicit}). In the next section, we will consider a simplified formal result, with a detailed proof.

\subsection{Simplified formal result}
In order to state a precise result, we will cast the flow on probability measures on $\mathcal{W} = \rb^{d+1}$ to a flow on measures on the unit sphere $$\mathcal{S}^d = \{ w \in \rb^{d+1}, \ \| w \|_2 = 1\}.$$
This is possible when the function $\Psi$ is positively $2$-homogeneous on $\mathcal{W} = \rb^{d+1}$, that is, such that $\Psi(\lambda w ) = \lambda^2 \Psi(w)$ for $\lambda > 0$. We can use homogeneity by reparameterizing each particle $w_j$ in polar coordinates as 
$$w_j = r_j \eta_j, \mbox{ with } r_j \in \rb \mbox{ and } \eta_j \in \mathcal{S}^d.$$
Using homogenetity, we have a prediction function:
$$
h =  \frac{1}{m} \sum_{j=1}^m \Psi(w_j) = \frac{1}{m} \sum_{j=1}^m r_j^2 \Psi( \eta_j).
$$
Moreover the function $J$ defined in \eq{J} is also $2$-homogeneous, and its gradient then $1$-homogeneous.
The flow from \eq{gf}, can be written
$$
\dot{w}_j = -\nabla J(w_j | \mu) \mbox{ with } \mu = \frac{1}{m} \sum_{i=1}^m \delta_{w_j}.
$$
A short calculation shows that the flow
\BEQ
\label{eq:fleta}
\left\{\begin{array}{ll}
\dot{r}_j & = - 2 r_j J(\eta_j | \nu ) \\
\dot{\eta}_j & = -  ( I - \eta_j \eta_j^\top) \nabla J(\eta_j|\nu) \\
\end{array} \right. \mbox{ with }  \nu = \frac{1}{m} \sum_{i=1}^m r_j^2 \delta_{\eta_j},  
\EEQ
leads to exactly the same dynamics. Indeed, by homogeneity of $\Psi$, the two definitions of $\mu$ and $\nu$ (through the $w_j$'s, or the $\eta_j$'s and $r_j$'s) lead to the same functions $J(\cdot|\mu)$ and $J(\cdot|\nu)$, and we get
\BEAS
\dot{w}_j & \ \ =\ \  &  \dot{r}_j \eta_j + r_j \dot{\eta}_j = - 2 r_j J(\eta_j | \nu ) \eta_j -  r_j ( I - \eta_j \eta_j^\top) \nabla J(\eta_j|\nu) \\
& \ \ =\ \  & - r_j  \nabla J(\eta_j|\nu) -  r_j \big[  2 J(\eta_j | \nu )  - \eta_j  ^\top \nabla J(\eta_j|\nu)  \big] \eta_j \\
& \ \ =\ \  &  - \nabla J(w_j|\mu),
\EEAS
because $w \mapsto \nabla J(w|\nu)$ is $1$-homogeneous and by the Euler identity for the $2$-homogeneous function 
 $w \mapsto   J(w|\nu) = J(w|\mu)$.
 
 Moreover, the flow defined in \eq{fleta} is such that $\eta_j$ remains on the sphere $\mathcal{S}^d$. We will study this flow under the assumption that the function $\Psi$ is sufficient regular, which excludes ReLU neural networks, but makes the proof easier (see more details in~\cite{chizat2021sparse}).
 
We first derive a PDE analogous to \eq{mu}. We consider a smooth test function $f: \mathcal{S}^d \to \rb$, and the quantity 
$$a = \int_{\mathcal{S}^d}  f(\eta) d\nu(\eta)= \frac{1}{m} \sum_{j=1}^m r_j^2 f(\eta_j).$$
We have
\BEA
\notag \dot{a} & \ \ =\ \  &\frac{1}{m} \sum_{j=1}^m 2 r_j \dot{r}_j f(\eta_j)
+ \frac{1}{m} \sum_{j=1}^m  r_j^2 \nabla f(\eta_j)^\top \dot{\eta}_j \\
\notag & \ \ =\ \  &- \frac{1}{m} \sum_{j=1}^m 4 r_j ^2   J(\eta_j | \nu )  f(\eta_j)
- \frac{1}{m} \sum_{j=1}^m  r_j^2 \nabla f(\eta_j)^\top ( I - \eta_j \eta_j^\top) \nabla J(\eta_j|\nu)\\
\label{eq:adot} & \ \ =\ \  & -  4 \int_{\mathcal{S}^d} f(\eta) J(\eta | \nu) d\nu(\eta)
-  \int_{\mathcal{S}^d} \nabla f(\eta|\nu)^\top ( I - \eta \eta^\top ) \nabla J(\eta | \nu) d\nu(\eta).
\EEA
This exactly shows that we have the PDE for the density $\nu_t$ at time $t$
\BEQ
\label{eq:nu}
\partial_t \nu_t(\eta) = - 4 J(\eta | \nu_t)  + \mathrm{div}( \nu_t(\eta) \nabla J(\eta | \nu_t) )
\EEQ
satisfied in the sense of distributions. We can now state our main result.
 
\begin{theorem}
\label{theo:2}
Assume the function $\Psi: \mathcal{S}^d \to \mathcal{F}$ is $d$-times continuously differentiable.
Assume $\nu_0$ is a nonnegative measure on the sphere $\mathcal{S}^d$ with finite mass and full support. Then the flow defined in \eq{nu} is well defined for all $t \geqslant 0$. Moreover, if $\nu_t$ converges weakly to some limit $\nu_\infty$, then $\nu_\infty$ is a global minimum of the function 
$\nu \mapsto F(\nu) = \displaystyle R \Big(  \int_{\mathcal{S}^d} \Psi(\eta) d\nu(\eta) \Big)$ over the set of nonnegative measures. 
\end{theorem}

\subsection{Proof of Theorem~\ref{theo:2}}
The global optimality conditions for minimizing the convex functional $F$ is that on the support of $\nu_\infty$ then $J(\eta|\nu_\infty)=0$, while on the entire sphere $J(\eta|\nu_\infty) \geqslant 0$.
The proof, adapted from~\cite{chizat2021sparse}, then goes a follows:
\BIT 
\item The existence and uniqueness of the flow $(\nu_t)_{t\geq 0}$ can be proved by using the equivalence with a Wasserstein gradient flow $(\mu_t)_{t\geq 0}$ in $\mathcal{P}(\rb^{d+1})$ and the theory of Wasserstein gradient flows~\cite{ambrosio2008gradient}. As a matter of fact, $(\nu_t)_{t\geq 0}$ is itself a gradient flow for a certain metric between nonnegative measures that is, in a certain sense, the inf-convolution between the Wasserstein and the Hellinger metric, see the discussion in~\cite{chizat2021sparse}.

\item The flow $\nu_t$ has a full support at all time $t$. This can be deduced from the representation of the solutions to Eq.~\eqref{eq:nu} as
$$
\nu_t =X(t,\cdot)_\# \Big(\nu_0 \exp\Big(-4 \int_0^t J( X(s,\cdot)| \nu_s ) ds\Big)\Big),
$$
where $X:[0,+\infty[\times \mathcal{S}^d\to\mathcal{S}^d$ is the flow associated to the time-dependent vector field $-\nabla J(\cdot \mid \nu_t)$, i.e., it satisfies $X(0,\eta)=\eta$ and $\frac{d}{dt} X(t,\eta) = -\nabla  J(X(t,\eta) | \nu_t)$ for all $\eta\in \mathcal{S}^d$, see, e.g.,~\cite{maniglia2007probabilistic}. Under our regularity assumptions, standard stability results for ODEs guarantee that at all time $t$, $X(t,\cdot)$ is a diffeomorphism of the sphere. Thus $\nu_t$ is the image measure (this is what the ``sharp'' notation stands for) by a diffeomorphism of a measure of the form $\nu_0 \exp(\dots)$ which has full support and thus $\nu_t$ has full support.

\item We assume that the flow converges to some measure $\nu_\infty$ (which could be singular). From \eq{fleta}, this imposes by stationarity of $\nu_\infty$ that
$J(\eta|\nu_\infty) = 0$ on the support of $\nu_\infty$, but nothing is imposed beyond the support of $\nu_\infty$ (and we need non-negativity of $J(\eta|\nu_\infty) $ for all $\eta \in \mathcal{S}^d$).

In order to show that $\min_{\eta \in \mathcal{S}^d} J(\eta|\nu_\infty) \geqslant 0$, we assume that it is strictly negative and will obtain a contradiction. We first need a $v < 0$ such that $v > \min_{\eta \in \mathcal{S}^d} J(\eta|\nu_\infty) $, and  the gradient $\nabla J(\eta|\nu_\infty)$ does not vanish on the $v$-level-set  $\{ \eta \in \mathcal{S}^d , J(\eta|\nu_\infty)=v\}$ of $J(\cdot | \nu_\infty)$. Such a $v$ exists because of Morse-Sard lemma which applies because under our assumptions, $J(\cdot|\nu)$ is $d$-times continuously differentiable for any finite nonnegative measure $\nu$.

We then consider the set $K = \big\{ \eta \in \mathcal{S}^d, \ J(\eta|\nu_\infty)  \leqslant   v \big\} $, which has some boundary $\partial K$, such that the gradient $\nabla J(\eta|\nu_\infty)$ has strictly positive dot-product with an outward normal vector to the level set at $\eta \in \partial K$. 

Since $\nu_t$ converges weakly to $\nu_\infty$, there exists $t_0>0$ such that for all $t\geqslant t_0$, $\sup_{\eta \in K} J(\eta|\nu_t) <   v/2$, while on the boundary $\nabla J(\eta|\nu_\infty)$ has non-negative dot-product with an outward normal vector. This means that for all $t>t_0$, applying \eq{adot} to the indicator function of $K$, if $a_t = \nu_t(K)$,
$$
a'(t) \geqslant - 4 \sup_{\eta \in K} J(\eta|\nu_t)  a(t).
$$
By the previous point, $a(t_0)>0$ and thus, by Gr\"onwall's lemma, $a(t)$ diverges, which is a contradiction with the convergence of $\nu_t$ to $\nu_\infty$.

\EIT

\subsection{Experiments}
\label{sec:experr}
\begin{figure}
\begin{center}
\includegraphics[scale=0.4,trim=0cm 0cm 0cm 0cm,clip]{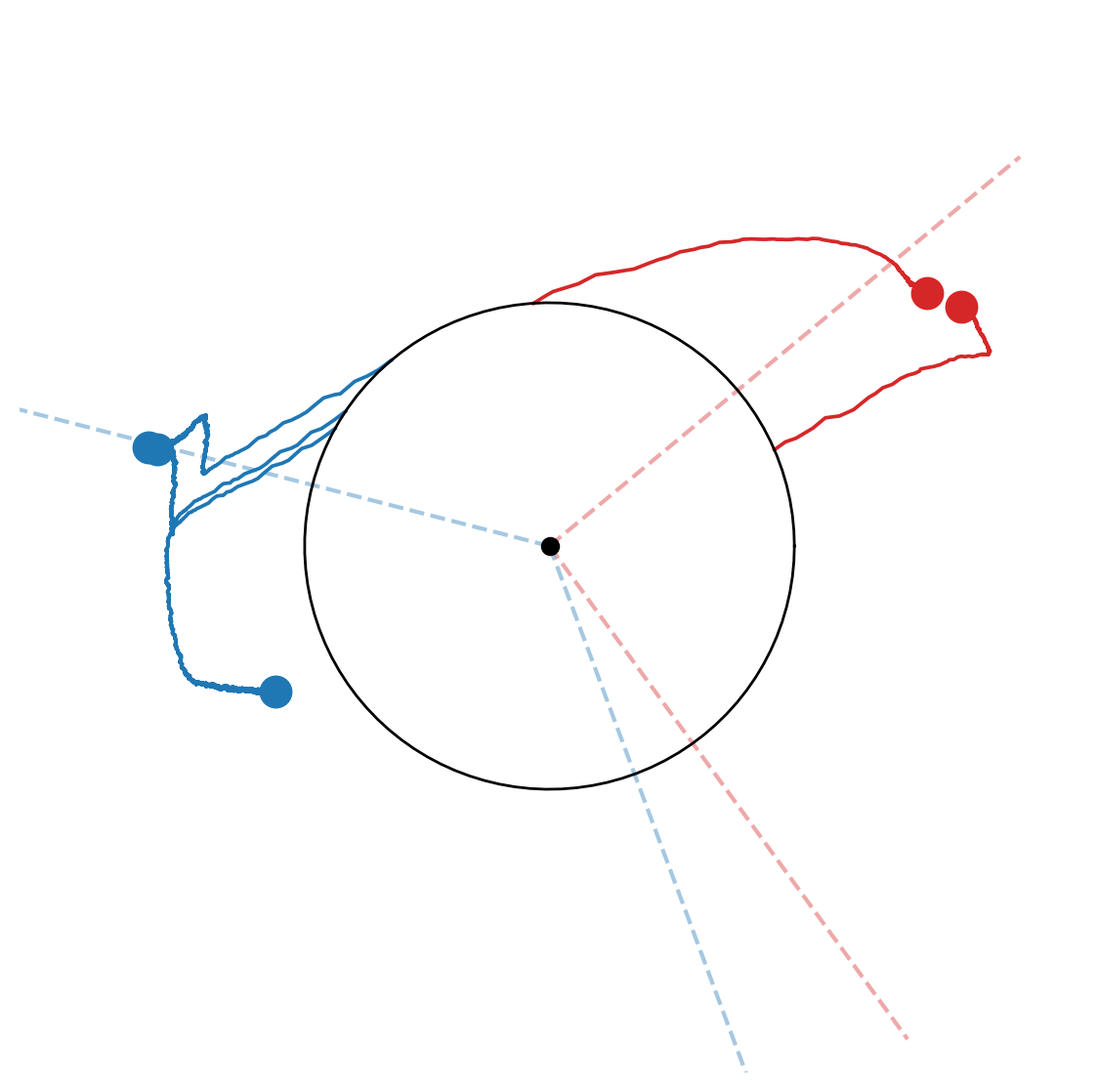}
\includegraphics[scale=0.4,trim=0cm 0cm 0cm 0cm,clip]{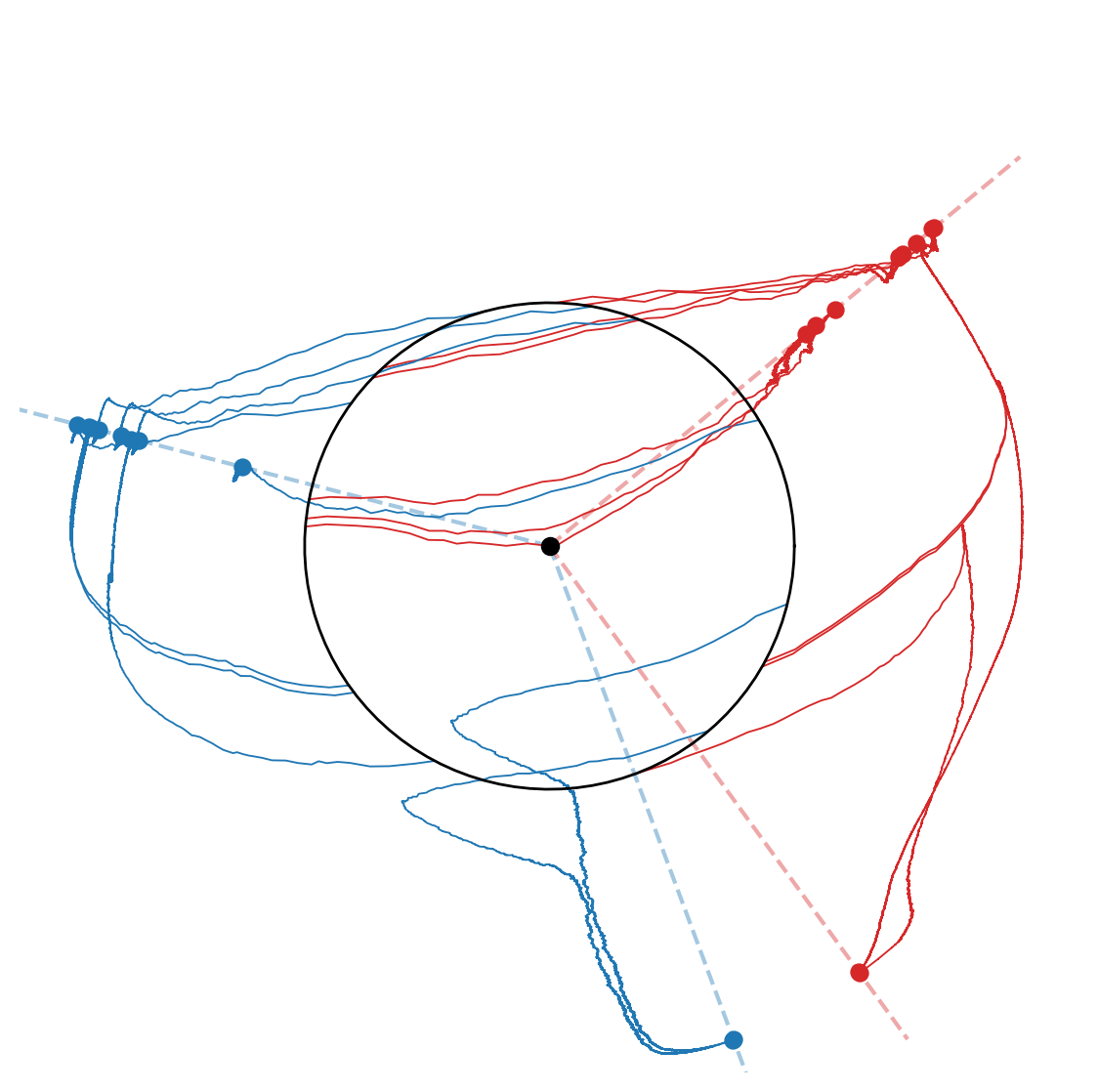}
\includegraphics[scale=0.4,trim=0cm 0cm 0cm 0cm,clip]{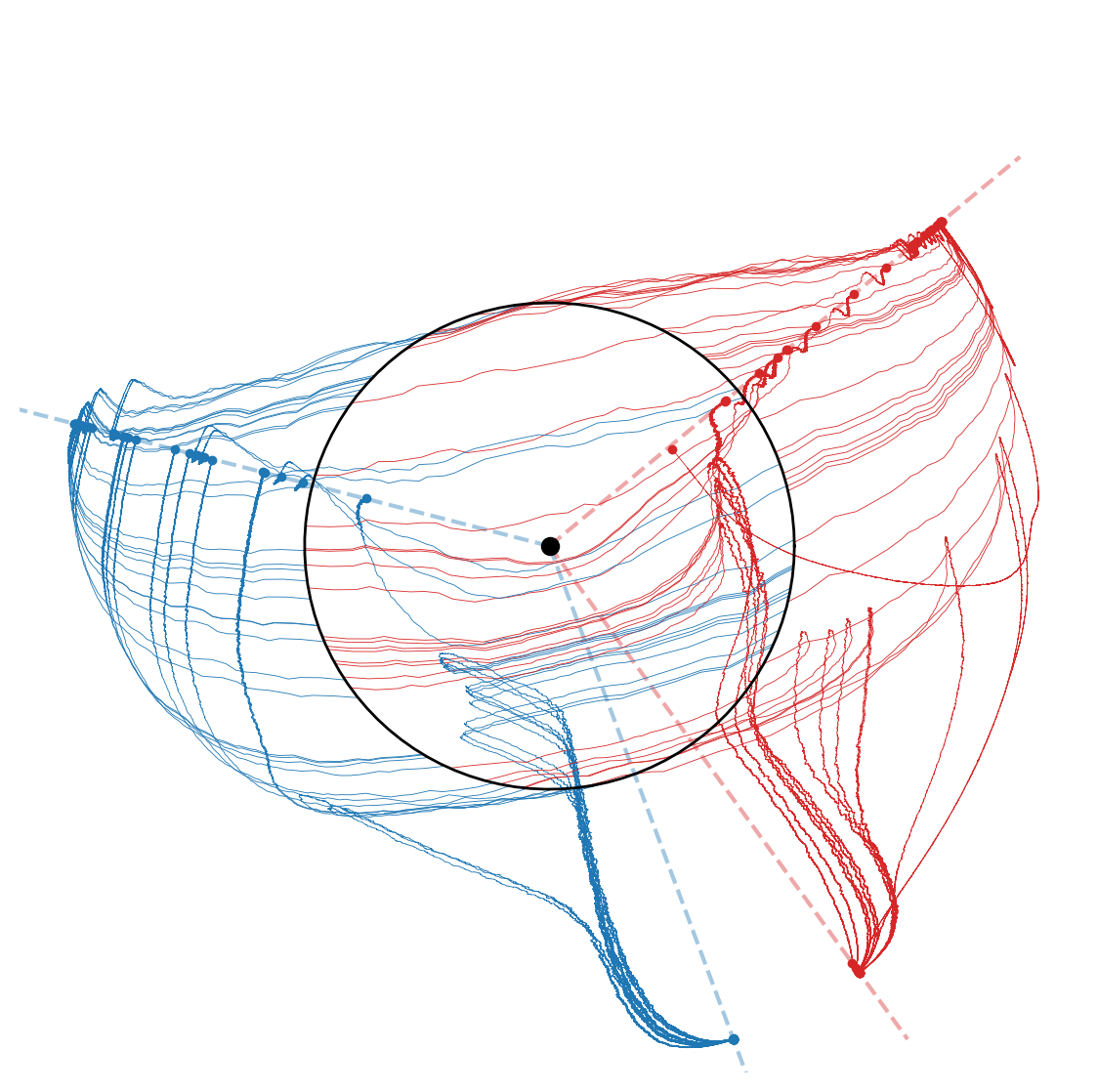}
\includegraphics[scale=0.4,trim=0cm 0cm 0cm 0cm,clip]{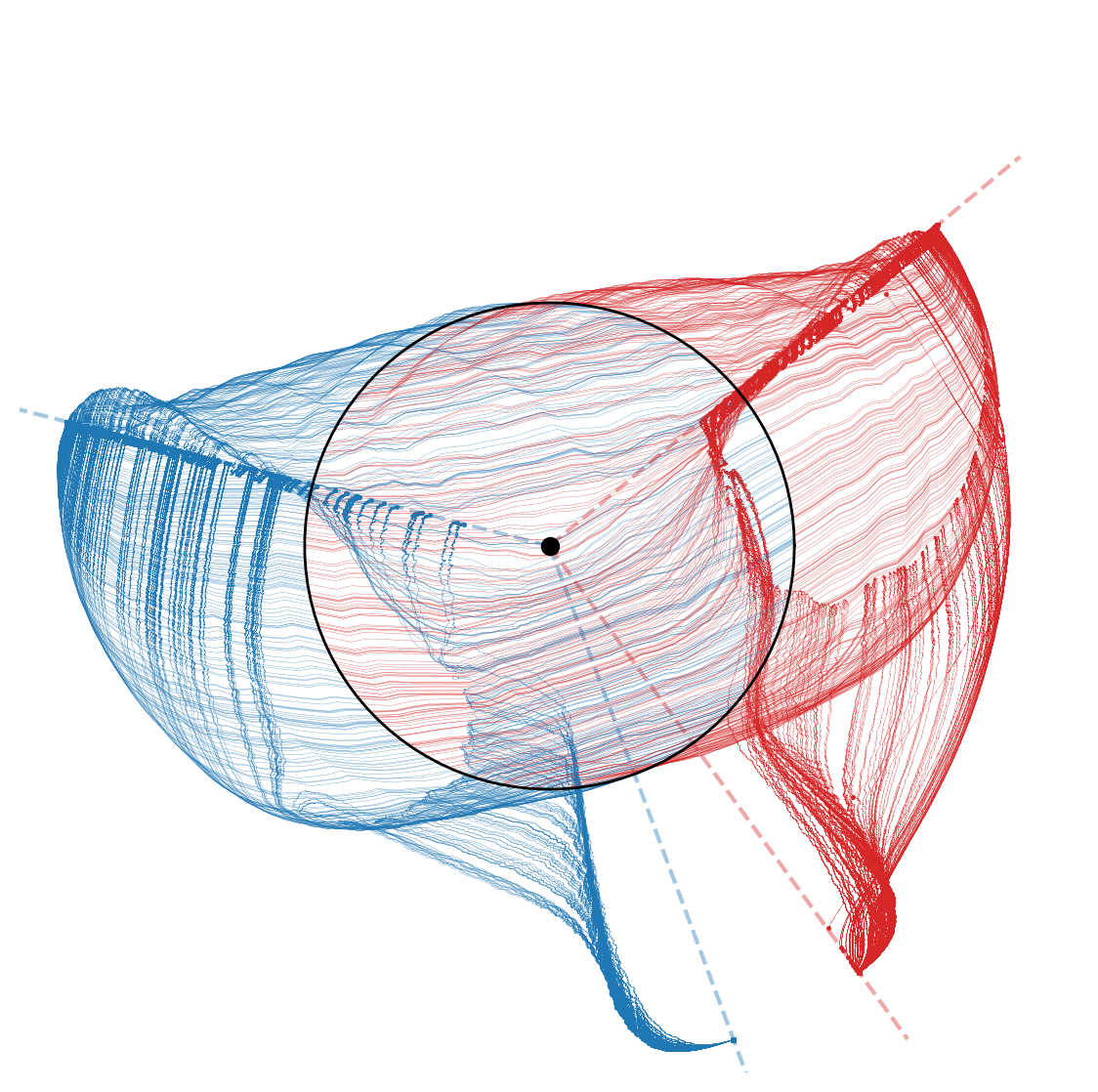}
\end{center}

\vspace*{-1cm}

\caption{Gradient flow on a two-layer ReLU neural network with respectively $m=5$, $m=20$, $m=100$ and $m=1000$. The position of the particles is given by $\vert \theta_2(j)\vert \cdot \theta_1(\cdot,j)$ and the color depends on the sign of $\theta_2(j)$. The dashed directions represent the neurons of the network that generates the data distribution (with $m_0=4$). The unit circle, where the particles are initialized, is plotted in black and the radial axis is scaled by $\tanh$ to improve lisibility.
\label{fig:nnplots} }
\end{figure}

\begin{figure}
\begin{center}
\includegraphics[scale=0.40,trim=0.2cm 0.2cm 0.2cm 0.2cm,clip]{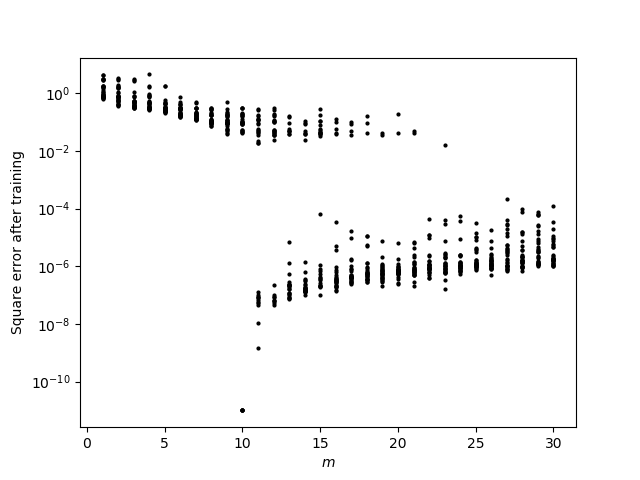}
\includegraphics[scale=0.40,trim=0.2cm 0.2cm 0.2cm 0.2cm,clip]{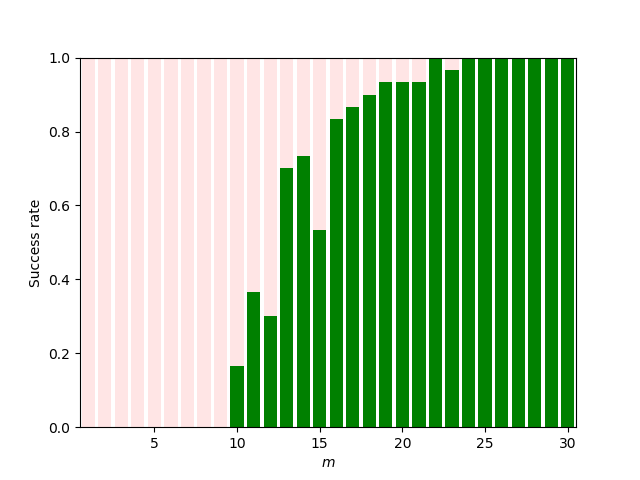}
\end{center}

\vspace*{-.8cm}

\caption{SGD on the square loss in the ``teacher-student'' setting ($10^4$ iterations, batch size $100$, learning rate $0.005$, $d=100$, the teacher has $m_0=10$ neurons). (left) Risk (expected square loss) after training  as a function of $m$ over $30$ random repetitions; (right) Success rate as a function of $m$  over $30$ repetitions (success means that the risk after training is below $10^{-3}$). \label{fig:teacher}}
\end{figure}

In order to illustrate\footnote{The code to reproduce Figures~\ref{fig:nnplots} and~\ref{fig:teacher} is available on this webpage~\url{https://github.com/lchizat/2021-exp-ICM}.} the global convergence result from earlier sections, we consider a supervised learning problem on $\rb^2$, with Gaussian input data $x$, and output data given by a ``teacher'' neural network
$$y = \sum_{j=1}^{m_0} \theta_2(j) \max\{ \theta_1(:,j)^\top x,0\}$$ for some finite $m_0$ and weights $\theta_1$ and $\theta_2$. We consider $R(h)$ the expected square loss and stochastic gradient descent with fresh new samples $(x_i,y_i)$ and a small step-size.

 We consider several number $m$ of hidden neurons, to assess when the original neurons can be recovered. In Figure~\ref{fig:nnplots}, for large $m$ (e.g., $m=100$ or $m=1000$), all learned neurons converge to the neurons that generated the function which is in accordance with our main global convergence result (note that in general, recovering the neurons of the teacher is not a necessary condition for optimality, but it is always sufficient), while for $m=5>m_0$, where the global optimum will lead to perfect estimation, we may not recover the global optimum with a gradient flow. An interesting open question is to characterize mathematically the case $m=20$, where we obtain the global optimum with moderate $m$.
 
In Figure~\ref{fig:teacher}, we consider several random initializations and random ``teacher'' networks and compute the generalization performance of the neural network after optimization. We see that for large $m$, good performance is achieved, while when $m$ is too small, local minima remain problematic. This experiment suggests that the probability of global convergence quickly tends to $1$ as $m$ increases beyond $m_0$ in this setting, even in moderately high dimension.

\section{Generalization guarantees and implicit bias for overparameterized models}
\label{sec:generalization}
A shown above, overparameterization -- which takes the form of large number of hidden neurons in our context -- is a blessing for optimization, as it allows to ensure convergence to a global minimizer. When stochastic gradient descent with fresh observations at each iteration is used, then the predictor will converge to the optimal predictor (that is, it will minimize the performance on unseen data), but will do so potentially at a slow speed, and with the need for many observations. In this context, overparameterization does not lead to overfitting, but may rather underfit.

In practice, several passes over a finite amount of data ($n$ observations) are used, and then overparameterization can in principle lead to overfitting. Indeed, among all predictors that will perfectly predict the training data, some will generalize, some will not. In this section, we show that the predictor obtained after convergence of the gradient flow can in certain cases be characterized precisely.

To obtain the simplest result, following~\cite{gunasekar2017implicit,soudry2018implicit,gunasekar2018characterizing} this will be done for binary classification problems with the logistic loss. We will first review the implicit bias for linear models before considering neural networks.

\subsection{Implicit bias for linear logistic regression}
In this section, we consider a linear model $h(x,\theta) = \theta^\top \Phi(x)$ and we consider the minimization of the unregularized empirical risk with the logistic loss, that is,
\BEQ
\label{eq:log}
\mathcal{R}(\theta) = \frac{1}{n} \sum_{i=1}^n \log \big( 1 + \exp( - y_i \theta^\top \Phi(x_i) \big).
\EEQ
We consider a \emph{separable} problem where there exists a linear function in $\Phi(x)$, $\theta^\top \Phi(x)$ such that $ y_i \theta^\top \Phi(x_i)  > 0$ for all $i \in \{1,\dots,n\}$. By rescaling, we may equivalently assume that there exists $\theta \in \rb^d$ such that
$$
\forall i \in \{1,\dots,n\}, \ y_i \theta^\top \Phi(x_i) \geqslant 1.
$$
This means that the objective function in \eq{log} has an infimal value of zero, which is not attained for any $\theta$, since it is strictly positive. However, taking any $\theta$ that separates the data as above, it holds that $\mathcal{R}(t\theta)$ converges towards $0$ as $t$ tends to infinity. There are thus in general an infinite number of directions towards which $\theta$ can tend to reach zero risk.

\begin{figure}[h]
\begin{center}
\includegraphics[width=6cm]{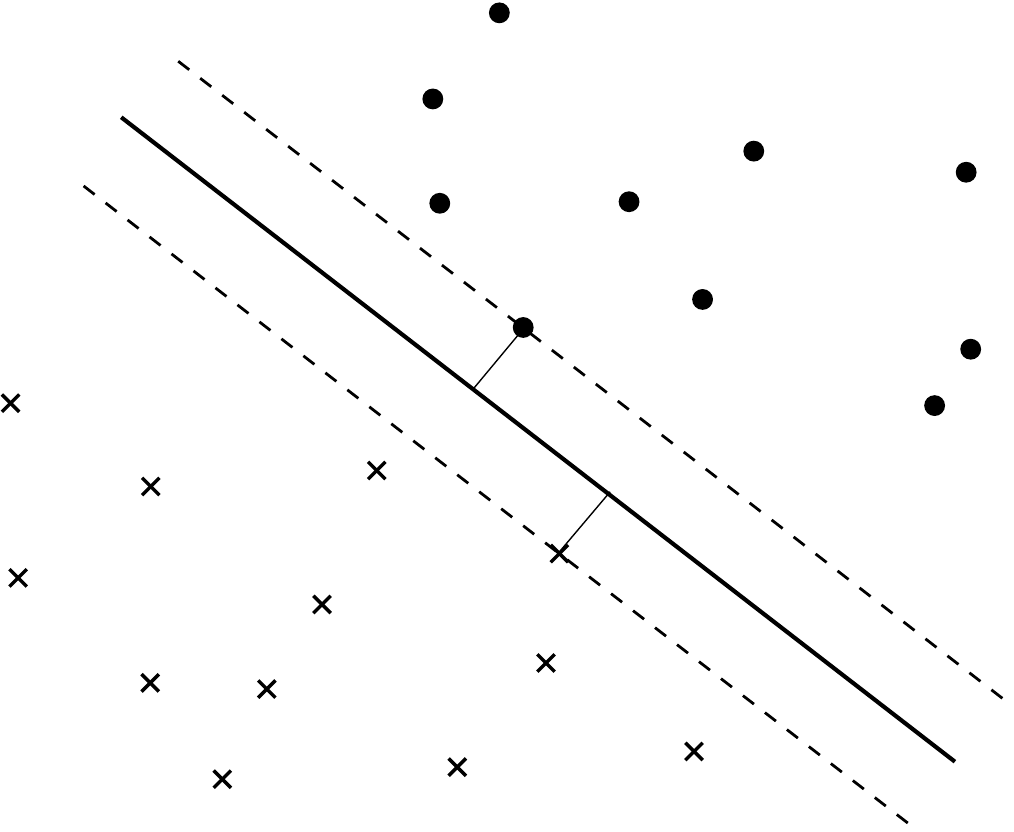} 
\end{center}

 \vspace*{-.5cm}
 
\caption{Geometric interpretation of \eq{mm} with a linearly separable binary classification problem in two dimensions (with each observation represented by one of the two labels  $\times$ or $\bullet$): among all separating hyperplanes going through zero, the one with the largest minimal distance from observations to the hyperplane will be selected. \label{fig:maxmarg}}
\end{figure}

It turns out that gradient descent selects a particular one: the iterate of gradient descent will diverge, but its direction (that is the element of the sphere it is proportional to) will converge \cite{soudry2018implicit} to the direction of a \emph{maximum margin classifier} defined as~\cite{svm1964} a solution to
\BEQ
\label{eq:mm}
\min_{\theta \in \rb^d} \ \  \| \theta \|_2^2 \quad\mbox{subject to}\quad \forall i \in \{1,\dots,n\}, \ y_i \theta^\top \Phi(x_i) \geqslant 1.
\EEQ
The optimization problem above has a nice geometric interpretation (see Figure~\ref{fig:maxmarg}). These classifiers with a large margin has been shown to have favorable generalization guarantees in a wide range of contexts~\cite{koltchinskii2002empirical}.

\subsection{Extension to two-layer neural networks}
We will now extend this convergence of gradient descent to a minimum norm classifier beyond linear models. We consider the minimization of the logistic loss
$$\frac{1}{n} \sum_{i=1}^n \log \big( 1 + \exp( - y_i h(x_i) \big),$$
where $h(x) = \frac{1}{m} \sum_{j=1}^m \theta_2(j) \max\{ \theta_1(:,j)^\top x,0\}$ is a two-layer neural network. We will consider two regimes: (1) where only the output weights $\theta_2(j)$, $j=1,\dots,m$ are optimized, and (2) where all weights are optimized. In these two situations, we will let the width $m$ go to infinity and consider the infinite-dimensional resulting flows. As shown in the previous section, when they converge, these flows converge to the global optimum of the objective function. But in the separable classification setting, the functions $h$ should diverge. We essentially characterize towards which directions they diverge, by identifying the norms that are implicitly minimized~\cite{chizat2020implicit}.

\subsection{Kernel regime}
\label{sec:kernel}
In this section, we consider random input weights $\theta_1(:,j)$, sampled from the uniform distribution on the sphere, and kept fixed throughout the optimization procedure. In other words, we only run the gradient flow with respect to the output weights $\theta_2 \in \rb^m$.

Since the model is a linear model with feature vectors in $m$ dimensions with components $$\Phi(x)_j = \frac{1}{\sqrt{m}} 
\max\{ \theta_1(:,j)^\top x,0\},$$ we can apply directly the result above from \cite{{soudry2018implicit}}, and the resulting classifier will minimize implicitly $\| \theta_2\|_2^2$, that is the direction of $\theta_2$ will tend to a maximum margin direction.

In order to study the situation when the number of features $m$ tends to infinity, it is classical within statistics and machine learning to consider the kernel function $\hat{k}: \rb^d \times \rb^d\to \rb $ defined as
$$
\hat{k}(x,x') = \Phi(x)^\top \Phi(x') = \frac{1}{m} \sum_{j=1}^m 
\max\{ \theta_1(:,j)^\top x,0\} \max\{ \theta_1(:,j)^\top x',0\}.
$$
When $m$ tends to infinity, the law of large number implies that $\hat{k}(x,x')$ tends to
$$
k(x,x') = \E \Big[
\max\{ \eta^\top x,0\} \max\{ \eta^\top x',0\}
\Big],
$$
for $\eta$ uniformly distributed on the sphere.

Thus, we should expect that in the overparameterized regime, the predictor behaves like predictors associated with the limiting kernel function~\cite{neal1995bayesian,rahimi2007random}. It turns out that the kernel $k$ can be computed in closed form~\cite{cho2009kernel}, and that the reproducing kernel Hilbert space (RKHS) functional norm $\Vert \cdot \Vert$ associated to the kernel $k$ is well understood (see below for a formula that defines it). In particular, this norm is infinite unless the function is at least $d/2$-times differentiable~\cite{bach2017breaking}, and thus very smooth in high dimension (this is to be contrasted with the fact that each individual neuron leads to a non-smooth function). We thus expect smooth decision boundaries at convergence (see experiments below). This leads to the following result (see details in~\cite{chizat2020implicit}):

\begin{theorem}[informal] When $m, t \to +\infty$ (limits can be interchanged), the predictor associated to the gradient flow converges (up to normalization) to the function in the RKHS that separates the data with minimum \emph{RKHS norm} $\Vert \cdot \Vert$, that is the solution to
$$
\min_{f} \ \Vert f\Vert^2 \quad\mbox{subject to}\quad \forall i \in \{1,\dots,n\}, \ y_i f(x_i) \geqslant 1.
$$
\end{theorem}

Note that the minimum RKHS norm function can also be found by using the finite-dimensional representation 
$f(x) = \sum_{i=1}^n \alpha_i k(x,x_i)$ and minimizing $\sum_{i,j=1}^n \alpha_i \alpha_j k(x_i,x_j)$ under the margin constraint, which is a finite-dimensional convex optimization problem.

A striking phenomenon is the absence of catastrophic overfitting, where the observed data are perfectly classified but with a very irregular function that would potentially not generalize well. Despite the strong overparameterization, the classifier selected by gradient descent can be shown to generalize through classical results from maximum margin estimation. See~\cite{mei2019generalization} for a related result where the performance as a function of $m$, and not only for infinite $m$, is considered in special settings. We will see a similar behavior when optimizing the two layers, but with a different functional norm.

\subsection{Feature learning regime}
We now consider the minimization with respect to both input and output weights. This will correspond to another functional norm that will not anymore be an RKHS norm, and will allow for more adaptivity, where the learned function can exhibit finer behaviors.

We first provide an alternative formulation of the RKHS norm as~\cite{bach2017breaking}
$$\displaystyle \| f\|^2 = \inf_{a(\cdot)} \int_{\mathcal{S}^{d-1}} |a(\eta)|^{\textcolor{red}{2}} d\tau(\eta) \ \mbox{ such that }\  f(x) = \int_{\mathcal{S}^{d-1}} ( \eta^\top x)_+ a(\eta) d\tau(\eta), $$ 
where the infimum is taken over all square-integrable functions on the sphere $\mathcal{S}^{d-1}$, and $\tau$ is the uniform probability measure on the sphere. This formulation highlights that functions in the RKHS combine infinitely many neurons.

We can then define the alternative \emph{variation norm}~\cite{kurkova} as
$$\displaystyle \Omega(f) = \inf_{a(\cdot)} \int_{\mathcal{S}^{d-1}} |a(\eta)| d\tau(\eta) \ \mbox{ such that }\  f(x) = \int_{\mathcal{S}^{d-1}} ( \eta^\top x)_+ a(\eta) d\tau(\eta), $$ where the infimum is now taken over all integrable functions on   $\mathcal{S}^{d-1}$. Going from squared $L_2$-norms to $L_1$-norms enlarges the space by adding non-smooth functions. For example, a single neuron corresponds to $a(\cdot) d \tau(\cdot)$ tending to a Dirac measure at a certain point, and thus has a finite variation norm.

This leads to the following result (see details and full set of assumptions in~\cite{chizat2020implicit}).

\begin{theorem}[informal] When $m,t \to +\infty$, if the predictor associated to the gradient flow converges (up to normalization), then the limit is the function that separates the data with minimum \textbf{variation norm} $\Omega(f)$, that is the solution to
$$
\min_{f} \ \Omega(f) \quad\mbox{subject to}\quad \forall i \in \{1,\dots,n\}, \ y_i f(x_i) \geqslant 1.
$$
\end{theorem}

Compared to the RKHS norm result, there is no known finite-dimensional convex optimization algorithms to efficiently obtain the minimum variation norm algorithm. Moreover, the choice of an $L_1$-norm has a sparsity-inducing effect, where the optimal $a(\cdot) d \tau(\cdot)$ will often corresponds to singular measure supported by a finite number of elements of the sphere. These elements can be seen as features learned by the algorithm: neural networks are considered as  methods that learn representations of the data, and we provide here a justification with a single hidden layer. Such feature learning can be shown to lead to improved prediction performance in a series of classical situations, such as when the optimal function only depends on a few of the $d$ original variables~\cite{bach2017breaking,chizat2020implicit}.


\subsection{Experiments}

In this section, we consider a large ReLU network with $m=1000$ hidden units, and compare the implicit bias and statistical performances of training both layers -- which leads to a max margin classifier with the variation norm -- versus the output layer -- which leads to max margin classifier in the RKHS norm. These experiments are reproduced from~\cite{chizat2020implicit}.

\begin{figure}\centering
\subfigure[Projected distribution]{\includegraphics[scale=0.43, trim = 0 -23 0 0, clip]{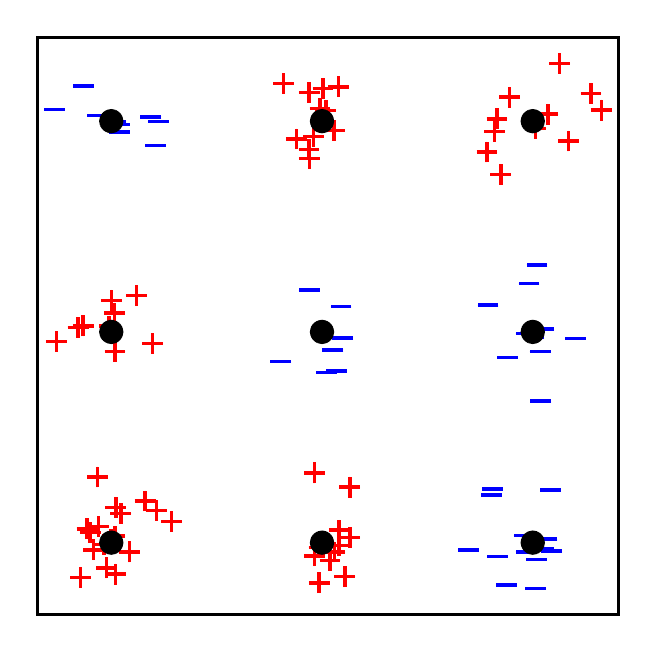}\label{fig:setting}}\hspace{0.01cm}
\subfigure[Test error vs.~$n$]{\includegraphics[scale=0.4]{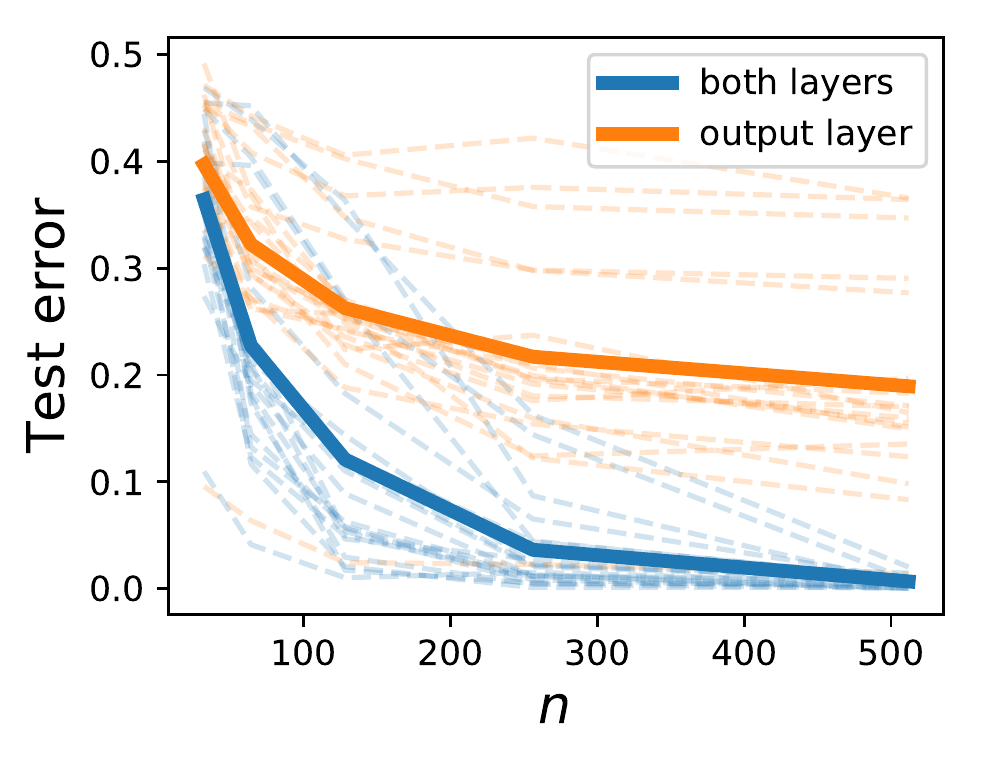}\label{fig:n}}\hspace{0.01cm}
\subfigure[Test error vs.~$d$]{\includegraphics[scale=0.4]{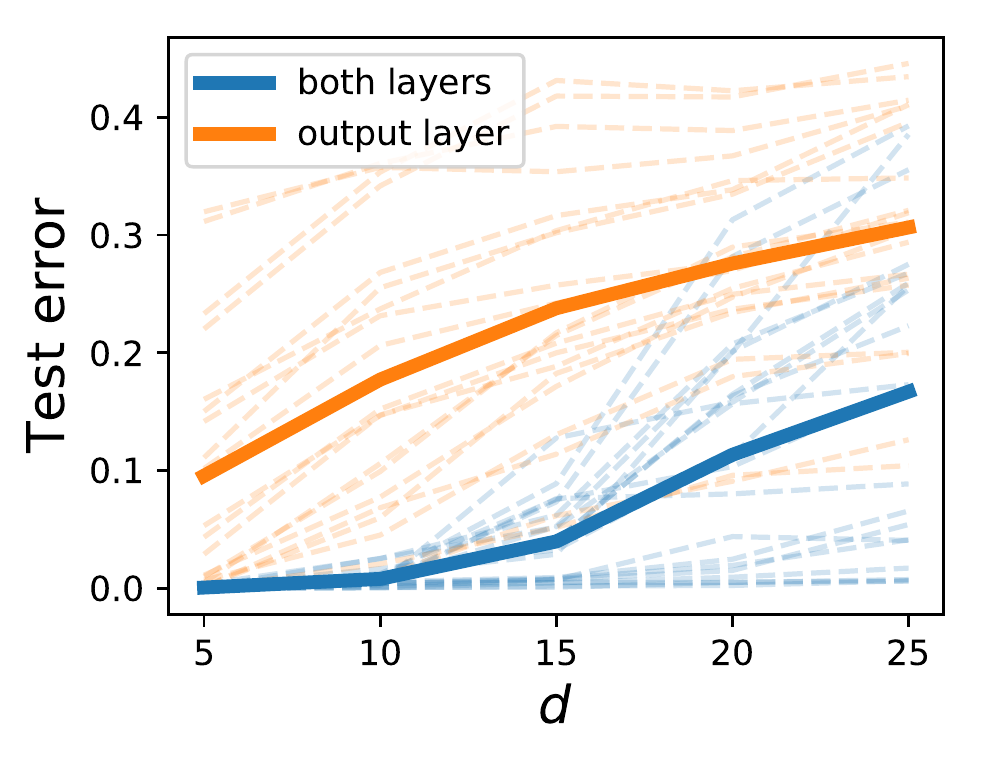}\label{fig:d}}\hspace{0.01cm}

\vspace*{-.25cm}

\caption{(a) Projection of the data distribution on the two first dimensions, (b) test error as a function of $n$ with $d=15$, (c) test error as a function of $d$ with $n=256$.}\label{fig:performance}
\end{figure}

\paragraph{Setting.}  Our data distribution is supported on $[-1/2, 1/2]^d$ and is generated as follows. In dimension $d=2$, the distribution of input variables is a mixture of $k^2$ uniform distributions on disks of radius $1/(3k-1)$ on a uniform $2$-dimensional grid with step $3/(3k-1)$, see Figure~\ref{fig:performance}(a) for an illustration with $k=3$. In dimension larger than $2$, all other coordinates follow a uniform distribution on $[-1/2,1/2]$. Each cluster is then randomly assigned a class in $\{-1,+1\}$.  

\begin{figure}
\centering
\begin{tabular}{rcccc}
\rotatebox{90}{$\quad$both layers}&\includegraphics[scale=0.18,trim=3.5cm 2cm 0.5cm 0,clip]{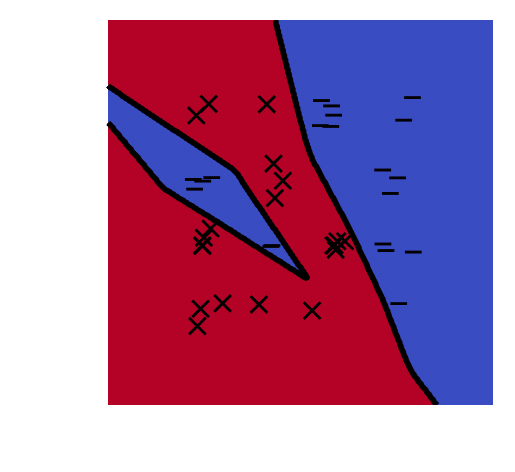} & \includegraphics[scale=0.18,trim=3.5cm 2cm 0.5cm 0,clip]{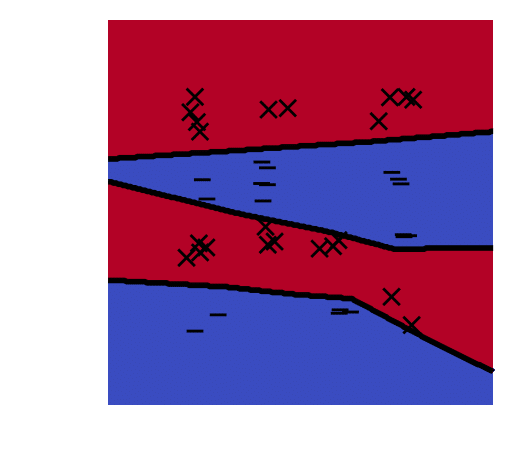}  & \includegraphics[scale=0.18,trim=3.5cm 2cm 0.5cm 0,clip]{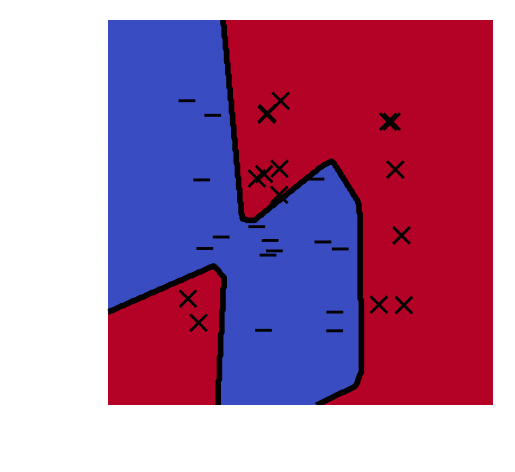}&  \includegraphics[scale=0.18,trim=3.5cm 2cm 0.5cm 0,clip]{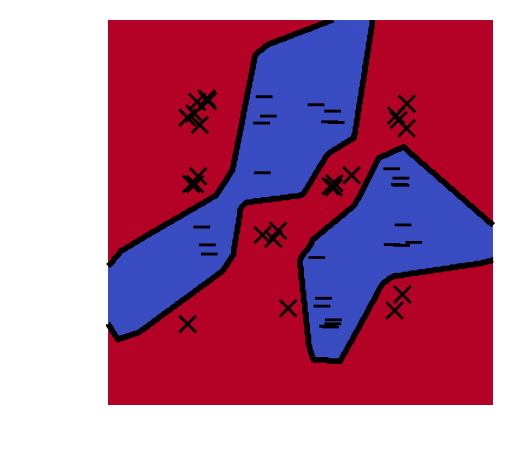} \\
\rotatebox{90}{$\;\;$output layer}&\includegraphics[scale=0.18,trim=3.5cm 2cm 0.5cm 0,clip]{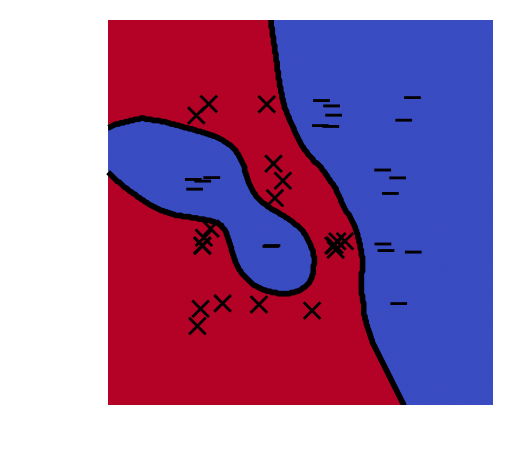} & \includegraphics[scale=0.18,trim=3.5cm 2cm 0.5cm 0,clip]{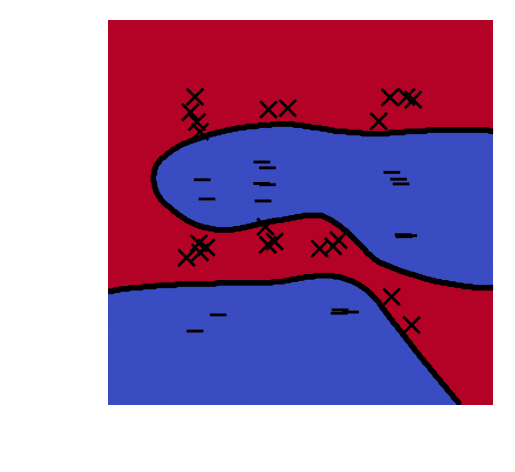}  &\includegraphics[scale=0.18,trim=3.5cm 2cm 0.5cm 0,clip]{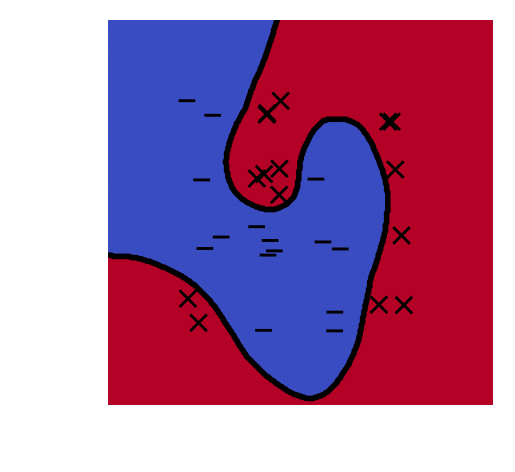}&  \includegraphics[scale=0.18,trim=3.5cm 2cm 0.5cm 0,clip]{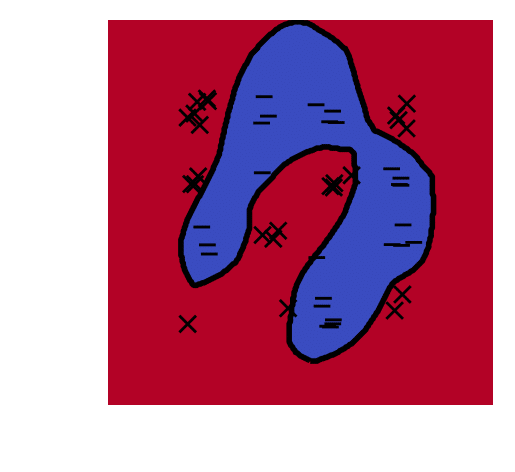}  \\
\end{tabular}
\vspace{-0.3cm}
\caption{Comparison of the implicit bias of training (top) both layers versus (bottom) only the output layer for wide two-layer ReLU networks with $d=2$ and for $4$ different random training sets.}\label{fig:illustration}
\end{figure}

\paragraph{Low dimensional illustrations.} Figure~\ref{fig:illustration} illustrates the differences in the implicit biases when $d=2$. It represents a sampled training set and the resulting decision boundary between the two classes for $4$ examples. The variation norm max-margin classifier  is non-smooth and piecewise affine, which comes from the fact that the $L_1$-norm favors sparse solutions. In contrast, the max-margin classifier for the RKHS norm has a smooth decision boundary, which is typical of learning in a RKHS.

\paragraph{Performance.} In higher dimensions, we observe the superiority of training both layers by plotting the test error versus $m$ or $d$ on Figure~\ref{fig:performance}(b) and~\ref{fig:performance}(c). We ran $20$ independent experiments with $k=3$ and show with a thick line the average of the test error $\mathbb{P}(yf(x)<0)$ after training.  For each $m$, we ran $30$ experiments using fresh random samples from the same data distribution.

\section{Discussion}
In this paper, we have presented qualitative convergence guarantees for infinitely-wide two layer neural networks. These were obtained with a precise scaling -- in the number of neurons -- of the prediction function, the initialization and the step-size used in the gradient flow. With those scalings, the mean-field limit exhibits feature learning capabilities, as illustrated in binary classification where precise functional spaces could be used to analyse where optimization converges to. However, this limit currently does not lead to quantitative guarantees regarding the number of neurons or the convergence time, and obtaining such guarantees remains an open problem. This is an active area of research with, in particular, recent results concerning the local convergence~\cite{zhou2021local,akiyama2021learnability, chizat2021sparse} or global convergence under strong assumption on the data~\cite{li2020learning}. Moreover, extending this analysis to more than a single hidden layer or convolutional networks remains difficult.

Different scalings lead to different behaviors~\cite{chizat2019lazy}. In particular, there is a scaling for which the limit behaves as a kernel method (even though all layers are trained, and not just the output layer) leading to another RKHS norm with a larger space than the one from \mysec{kernel}, see~\cite{jacot2018neural}. While not leading to representation learning, extensions to deeper networks are possible with this scaling and provide one of few optimization and statistical guarantees for these models. Some recent progress has been made in the categorization of the various possible scalings for deep networks~\cite{yang2020feature}, and this emerging general picture calls for a large theoretical effort to understand the asymptotic behaviors of wide neural networks.

\label{sec:conclusion}


\paragraph{Acknowledgements.}
This work was funded in part by the French government under management of Agence Nationale de la Recherche as part of the ``Investissements d'avenir'' program, reference ANR-19-P3IA-0001 (PRAIRIE 3IA Institute). We also acknowledge 
 support the European Research Council (grant SEQUOIA 724063).


\bibliographystyle{emss} 

\bibliography{icm2022}

\end{document}